\documentclass[letterpaper]{article} 
\usepackage{aaai25}  
\usepackage{times}  
\usepackage{helvet}  
\usepackage{courier}  
\usepackage[hyphens]{url}  
\usepackage{graphicx} 
\usepackage{natbib}  
\usepackage{caption} 
\usepackage{algorithm}
\usepackage{algorithmic}

\usepackage{newfloat}
\usepackage{listings}
\DeclareCaptionStyle{ruled}{labelfont=normalfont,labelsep=colon,strut=off} 
\floatstyle{ruled}
\newfloat{listing}{tb}{lst}{}
\floatname{listing}{Listing}
\title{Learning Causal Transition Matrix for Instance-dependent Label Noise}
\author{
    Jiahui Li\textsuperscript{\rm 1}\textsuperscript{\rm 2}\equalcontrib, Tai-Wei Chang\textsuperscript{\rm 2}\equalcontrib, Kun Kuang\textsuperscript{\rm 1}\thanks{Corresponding Author},
    Ximing Li\textsuperscript{\rm 2}, Long Chen\textsuperscript{\rm 3}, Jun Zhou\textsuperscript{\rm 2} \\
}
\affiliations{
    \textsuperscript{\rm 1}Zhejiang University,
    \textsuperscript{\rm 2}Ant Group,
    \textsuperscript{\rm 3}The Hong Kong University of Science and Technology \\

    jiahuil@zju.edu.cn, taiwei.twc@antgroup.com, kunkuang@zju.edu.cn \\
    xili.lxm@antgroup.com, longchen@ust.hk, jun.zhoujun@antgroup.com \\
}

\usepackage{graphicx}
\usepackage{booktabs}

\usepackage[accsupp]{axessibility}  

\newcommand{\ie}{\textit{i}.\textit{e}.}
\newcommand{\eg}{\textit{e}.\textit{g}.}
\newcommand{\bigCI}{\mathrel{\text{\scalebox{1.07}{$\perp\mkern-10mu\perp$}}}}
\newtheorem{theorem}{Theorem}

\usepackage{multirow}
\usepackage{subcaption}
\usepackage{bibentry}
\begin{document}

\maketitle
\begin{abstract}
Noisy labels are both inevitable and problematic in machine learning methods, as they negatively impact models' generalization ability by causing overfitting. In the context of learning with noise, the \textbf{transition matrix} plays a crucial role in the design of statistically consistent algorithms. However, the transition matrix is often considered unidentifiable.
One strand of methods typically addresses this problem by assuming that the transition matrix is \textbf{instance-independent}; that is, the probability of mislabeling a particular instance is not influenced by its characteristics or attributes. This assumption is clearly invalid in complex real-world scenarios.
To better understand the transition relationship and relax this assumption, we propose to study the data generation process of noisy labels from a causal perspective. We discover that an unobservable latent variable can affect either the instance itself, the label annotation procedure, or both, which complicates the identification of the transition matrix.
To address various scenarios, we have unified these observations within a new causal graph. In this graph, the input instance is divided into a \textbf{noise-resistant component} and a \textbf{noise-sensitive component} based on whether they are affected by the latent variable. These two components contribute to identifying the ``causal transition matrix'', which approximates the true transition matrix with theoretical guarantee.
In line with this, we have designed a novel training framework that explicitly models this causal relationship and, as a result, achieves a more accurate model for inferring the clean label.

\end{abstract}

%

\section{Introduction}
\label{sec:intro}

The success of deep neural networks is heavily based on large-scale annotation datasets~\cite{daniely2019generalization,yao2020searching}. 
However, data annotation inevitably introduces label noise, and the models are prone to overfitting on the mislabeled data, which hampers their performance in terms of generalization. 
Cleaning up corrupted labels is extremely expensive and time-consuming.
Therefore, finding effective algorithms to mitigate the impact of label noise and improve model performance is of utmost importance. 

\begin{figure}[t]
\centering
\includegraphics[width=0.95\linewidth]{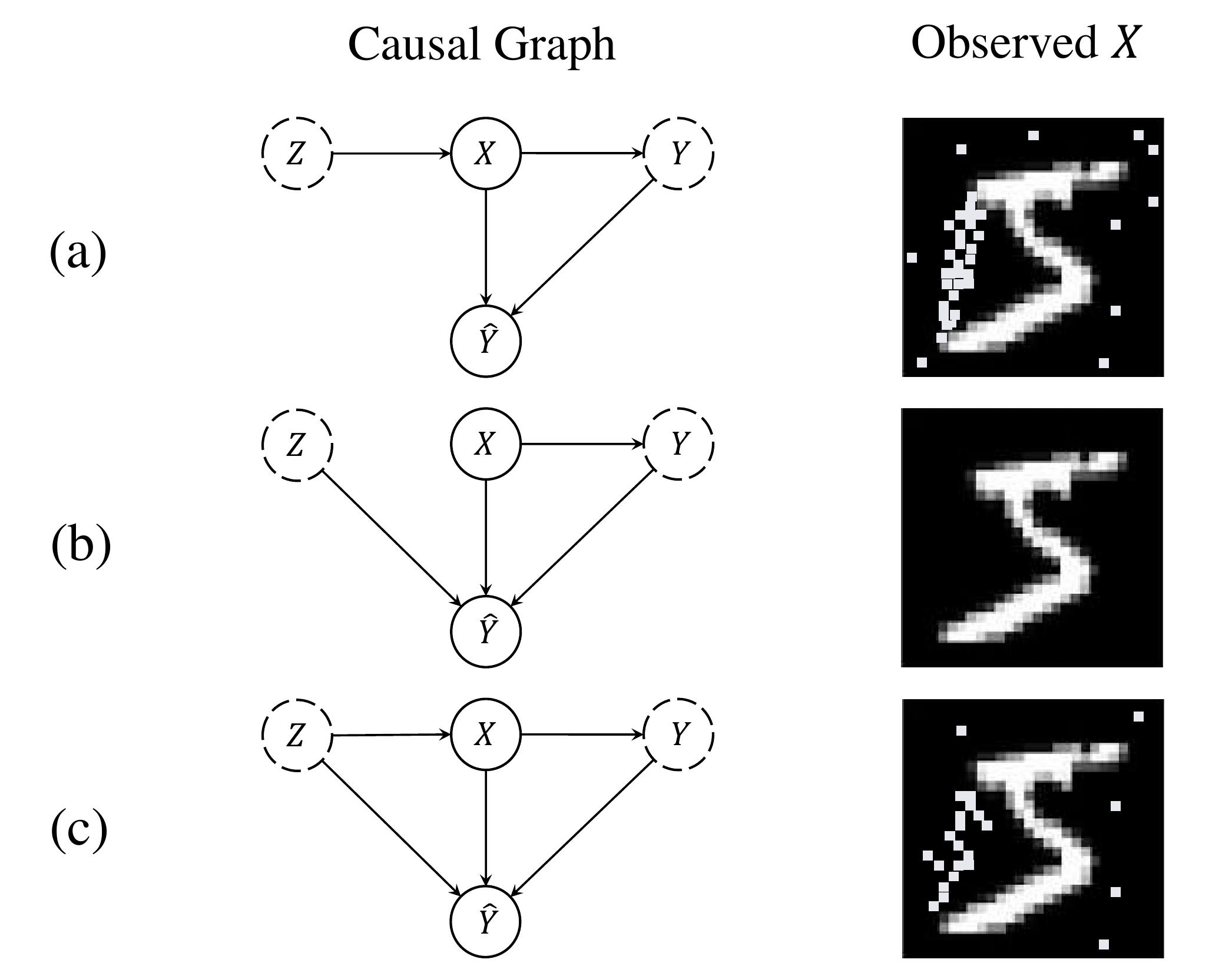}
  \caption{Examples of three causal graphs illustrating the mislabeling of ``5'' as ``6'' in MNIST, where $X$ denotes instance(image), $Y$ denotes the ground truth label, $\hat{Y}$ denotes the noisy label, and $Z$ denotes the latent variable. The dashed circles represent the unobservable variable. (a) The instance is perturbed by noise, making ``5'' looks like ``6''. (b) The instance is clean, but it is mislabeled by an annotator. (c) The instance exhibits a mixture situation of (a) and (b). }
  \label{fig:intro}
  \vspace{-1em}
\end{figure}

Most of the main research~\cite{patrini2017making,vahdat2017toward,xiao2015learning,cheng2020learning,yao2021instance,bae2022noisy,zhang2024cognition} in label-noise learning focuses on estimating the transition matrix $T_{ij}=P(\hat{Y}=j|Y=i,X=x)$, which captures the relationship between the true (clean) labels $Y$ and the observed (noisy) labels $\hat{Y}$. 
This transition matrix provides valuable information about the noise distribution and can be used to guide the learning process.
Typically, the identification of the transition matrix becomes a challenge due to the inaccessibility of the clean label $Y$ for direct observation.
In order to facilitate learning in the presence of noise, one strand of methods~\cite{natarajan2013learning,berthon2021confidence,xia2020part} assumes that the transition relationship between the noisy label $\hat{Y}$ and the clean label $Y$ is \textbf{instant-independent}, wherein $P(\hat{Y}|Y,X)=P(\hat{Y}|Y)$.
However, in complex real-world scenarios, label noise can be attributed to various factors and may violate the instance-independent assumption. Therefore, it is crucial to find a more rational and effective transition relationship to handle label noise in realistic and complex settings.

To explore the factors that hinder the inference of the clean label, we approach our study from a causal perspective.
Depending on the source of noise, we can get three distinct causal graphs of the data generation procedure, which are depicted in Figure~\ref{fig:intro}.
Firstly, in scenario (a), certain environmental factors $Z$, \eg lighting, noise, shadow, impact the instance $X$. This, in turn, influences the annotators, leading to the production of a noisy label $\hat{Y}$. Secondly, in scenario (b), some factors $Z$ do not influence $X$ but directly cause the generation of the noisy label $\hat{Y}$, such as the annotator's negligence.
Lastly, both scenarios (a) and (b) can occur simultaneously, resulting in a noisy label $\hat{Y}$.
In all the cases shown in the figure, fulfilling the instance-independent assumption is challenging. 
It is evident that $X$ serves as a common cause for both $Y$ and $\hat{Y}$, introducing confounding bias when attempting to estimate the transition matrix $P(\hat{Y} | Y, X)$. 
Furthermore, the presence of an unobservable latent variable $Z$ further complicates the estimation process, resulting in inaccuracies in learning the clean labels.

To address these challenges and improve the understanding of the transition relationship, we have integrated the three causal graphs presented in Figure~\ref{fig:intro} into a unified diagram, shown in Figure~\ref{fig:intro}(c), and we introduce a new causal graph in Figure~\ref{fig:method}(a). In this proposed causal graph, we consider $Y$ as the treatment and $\hat{Y}$ as the outcome and introduce a novel concept: the ``causal transition matrix'' $P(\hat{Y}|do(Y),X)$, which serves to approximate the true transition matrix.
The proposed causal graph suggests the presence of two components within $X$: a \textbf{noise-resistant component} $X_1$ and a \textbf{noise-sensitive component} $X_2$. The noise-resistant component $X_1$ is not influenced by the latent variable $Z$ but plays a role in determining the clean label $Y$. In contrast, the noise-sensitive component $X_2$ is subject to interaction with both the latent variable $Z$ and the clean label $Y$, contributing to the generation of the noisy label $\hat{Y}$.
The clear distinction of these two components enables the ``causal transition matrix`` $P(\hat{Y}|do(Y),X)$ to be identified with a theoretical guarantee, which in turn contributes to the inference of the clean label $Y$ from $X$.

Building upon this, we have developed a novel training framework that explicitly models the variables and causal relationships within the proposed causal graph. Initially, the framework separates the noise-resistant component $X_1$ and the noise-sensitive component $X_2$ from the variable $X$. Given that the variable $Y$ is unobservable, we use confidence sampling techniques to make it partially observable. In addition, we have designed a transition model to capture the relationships among various variables effectively. Our design ensures that the causal relationship is accurately established, leading to more robust and precise learning outcomes in the presence of label noise.

Our contribution can be summarized as follows:

(1) We propose a novel causal graph that integrates multiple scenarios in label-noise learning, thus facilitating a more comprehensive understanding of the transition dynamics in label-noise learning.

(2) Based on the causal graph, we relax the instance-independent assumption and introduce the concept of the ``causal transition matrix'' $P(\hat{Y}|do(Y),X)$, which serves as an approximation for the transition matrix and provides a theoretical guarantee for its identification.

(3) We propose a novel framework that explicitly models the variables and their causal relationships as delineated in the proposed causal graph. This framework can be trained end-to-end, leading to a more precise model to infer clean labels.

(4) Experiments on both synthetic and real-world label-noise datasets highlight the superiority of our method.






\section{Related Work}
Due to the susceptibility of the traditional cross entropy (CE) loss to overfit noisy labels, and the high cost and time involved in cleaning up corrupted labels, there has been a notable surge in interest in learning with noisy labels.

\paragraph{Transition matrix in label-noise learning.}
The transition matrix plays a crucial role in the learning of label noise as it captures the relationships between true labels and observed noisy labels. 
One line of approaches~\cite{patrini2017making,vahdat2017toward,xiao2015learning,cheng2020learning} involves developing algorithms that initially estimate the transition matrix roughly and then correct it using posteriors. 
However, recent studies~\cite{cheng2020learning,xia2019anchor,yao2020dual} have revealed that the transition matrix is unidentifiable and thus challenging to learn. 
To overcome this challenge, researchers~\cite{natarajan2013learning,berthon2021confidence,xia2020part} often assume that the noisy label is independent of the instance, denoted $P(\hat{Y}|Y,X) = P(\hat{Y}|Y)$. 
Additionally, some approaches~\cite{zhang2018generalized,wang2019symmetric,yao2020searching,ma2020normalized} focus on designing robust loss functions that mitigate the label noise issue without the need for estimating the transition matrix. However, as pointed out by \citet{zhang2021learning}, these methods may only perform adequately under simple conditions, such as symmetric noise, and may struggle when dealing with heavy and complex noise.
Another line of research~\cite{han2018co,han2020sigua,yu2019does,mirzasoleiman2020coresets,wu2020topological} involves performing sample selection during training, some of which can also eliminate the instance-independent assumption. 
These methods utilize techniques such as learning with rejection~\cite{el2010foundations,thulasidasan2019combating,mozannar2020consistent,charoenphakdee2021classification}, meta-learning~\cite{shu2019meta,li2019learning}, contrastive learning~\cite{li2022selective}, or semi-supervised learning~\cite{nguyen2019self,li2020dividemix} to select training samples with high confidence. 
However, as argued by \citet{cheng2020learning}, these methods cannot guarantee the avoidance of overfitting to label noise because they are trained based on CE loss.

\paragraph{Causal methods for label-noise learning.}
The causal perspective serves as a valuable tool to improve our understanding of the noise generation process. 
\citet{yao2021instance} propose CausalNL which uses variational inference~\cite{kingma2013auto} to model the causal structure underlying the generation of noisy data. 
Likewise, \citet{bae2022noisy} propose a noisy prediction calibration method based on the causal structure, which also adopts the generative model to explicitly model the relationship between the output of a classifier and the true label.
However, both rely on the assumption of instance independence in the posterior probability, $P(Y|\hat{Y},X)=P(Y|\hat{Y})$, and model such a relation via the variational auto-encoder~\cite{kingma2013auto}.


\noindent\textbf{Compared with the other causal-based methods.}
To establish a more accurate and reliable transition relationship, we propose a novel causal structure and framework for learning with noise. This approach does not require the use of variational inference tools and eliminates the need for the posterior probability assumption. Within this structure, we introduce the concept of ``causal transition matrix'', which approximates the true transition matrix with theoretical guarantees, thus enhancing the efficacy of learning under noise conditions.




\section{Method}
\subsection{Problem Setup}
Considering a classification problem with $k$ classes.
Let $X$ and $Y$ be the random variables for the input instance and the clean label respectively.
The training examples ${(x_n,y_n)}$ are sampled from the joint probability distribution $P$ over the random variables $(X,Y) \in \mathcal{X} \times \mathcal{Y}$, where $n \in \{1,2,...,N\}$ and $N$ represent the number of training examples. 
The goal of the classification task is to find a classifier $f: \mathcal{X} \rightarrow \mathcal{Y}$ that accurately maps $X$ to $Y$. 
Mainstream methods train the neural network by minimizing the empirical risk $\mathbb{E}_{P} \left[\mathcal{L}(f(x),y)\right]$, where $\mathcal{L}$ denotes the loss function. 
In real-world applications, we can only observe the sample $(x_n,\hat{y}_n)$, where $\hat{y}_n$ represents the noisy label. 
The noise of the label of each instance is characterized by a transition matrix $T(X)$, where each element $T_{ij}(X) =P(\hat{Y}=j|Y=i,X=x)$ denotes a probability from $Y$ to $\hat{Y}$ when given $X$. 
When learning with noisy labels, the loss function becomes $\mathcal{L}(f(x),\hat{y})$, but our objective is still to train a clean classifier that minimizes the empirical risk on the clean label $y$.

\subsection{Causal Viewpoint for Denoising}
\begin{figure*}[t]
    \centering
    \begin{subfigure}{0.22\linewidth}
     \centering
        \includegraphics[width=\linewidth]{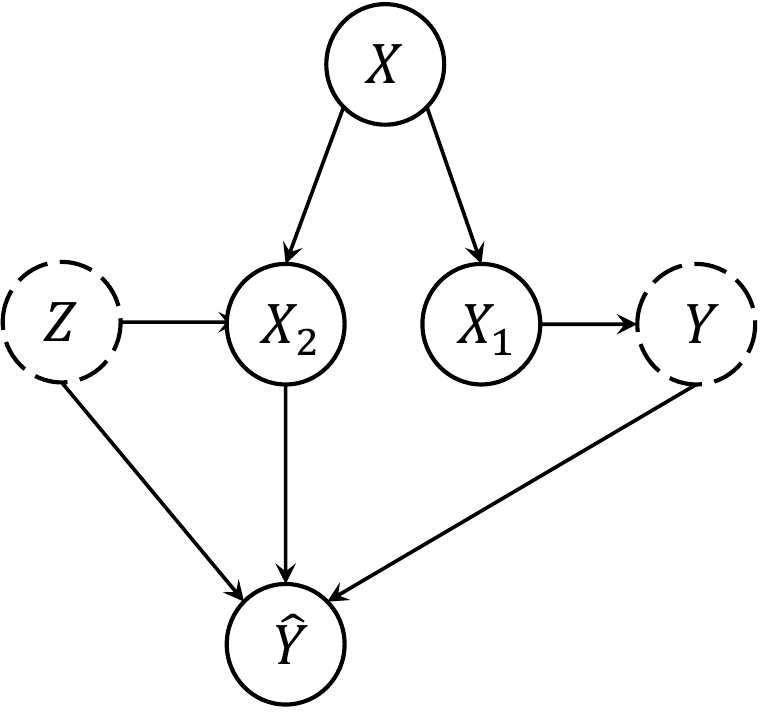}
        \caption{}
    \end{subfigure}
    \hspace{3mm}
    \centering
    \begin{subfigure}{0.55\linewidth}

      \includegraphics[width=\linewidth]{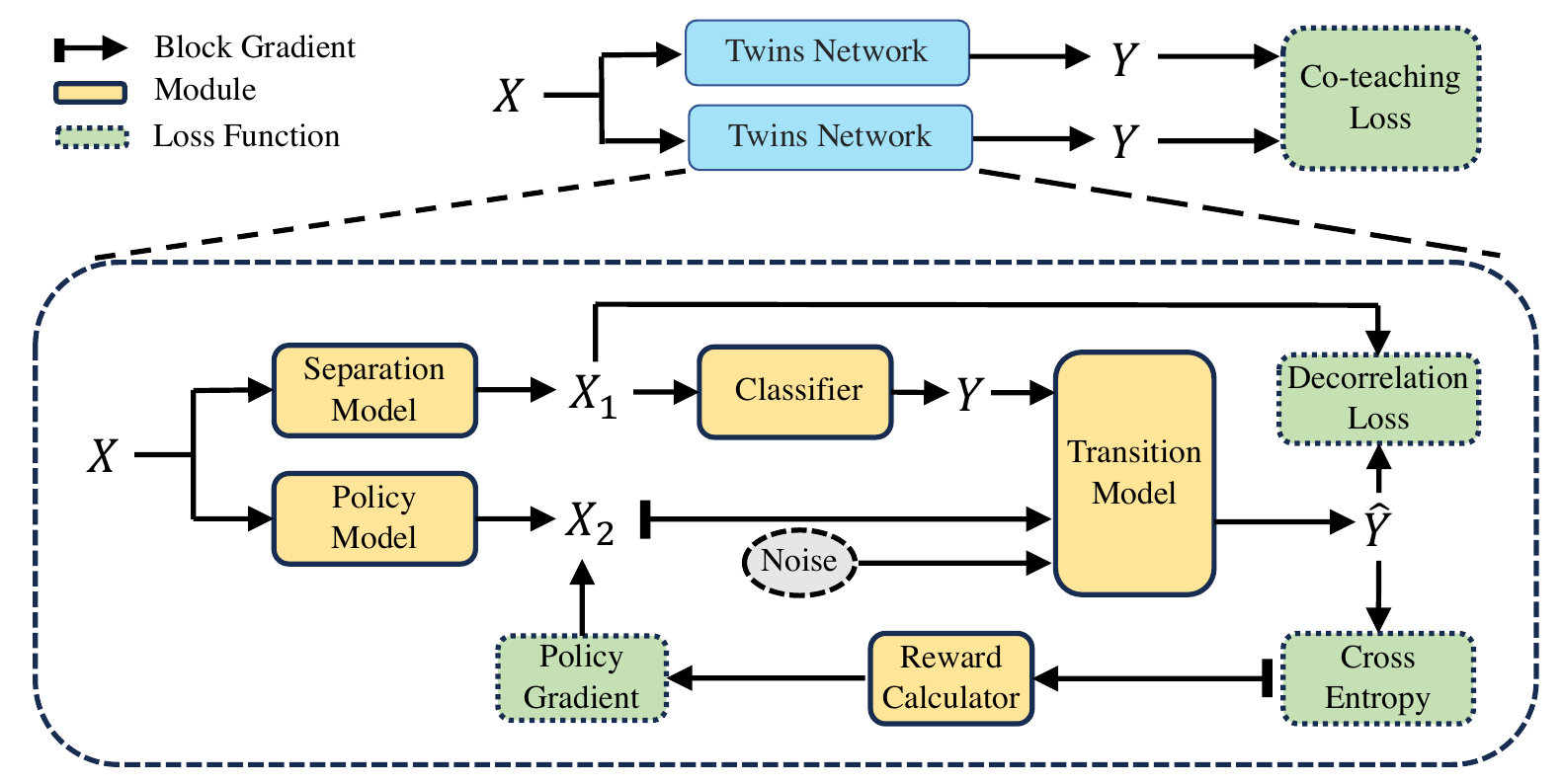}
    \caption{}    
    \end{subfigure}
    \caption{(a) The proposed causal graph for learning with noisy labels. (b) The training framework of our method. }
    \label{fig:method}
\end{figure*}

We have summarized the causal relationship in Figure~\ref{fig:method}(a), which involves two unobservable variables: $Z$ and $Y$. 
In recent literature~\cite{han2018co,han2020sigua,yu2019does,yao2021instance,bae2022noisy}, $\hat{Y}$ from reliable examples can be considered the clean label, which makes $Y$ partially observed. Consequently, the unobservable latent variable $Z$ becomes the primary obstacle.
Mainstream studies~\cite{cheng2020learning,xia2019anchor,yao2020dual} typically deem that the transition matrix $T(X)$ is not identifiable and difficult to learn. 
We introduce the concept of ``causal transition matrix'', which is defined as $T_{cau} = P(\hat{Y}|do(Y),X)$. 
This represents the outcome distribution of the noisy label $\hat{Y}$ by intervening on $Y$ conditioned on $X$. 
By intervening on $Y$, we essentially eliminate the incoming edge of $Y$, thus ensuring unbiased estimation.

In causal theory, a natural approach to estimate $T_{cau} = P(\hat{Y}|do(Y),X)$ is backdoor adjustment~\cite{pearl1995causal}, since $X$ effectively blocks all backdoor paths. However, accurately computing an unbiased relationship necessitates sampling across $X$, which introduces substantial computational challenges. Moreover, computing over $X$ is more ``noisy'', as it involves a lot of irrelevant information.  Consequently, an accurate and efficient alternative approach is necessary to investigate.

To make the causal transition matrix identifiable, we separate each instance $X$ into two components, the noise-resistant component $X_1$  and the noise-sensitive component $X_2$. 
We posit that $X_1$ is not influenced by the latent variable $Z$ and directly causes the clean label $Y$, while $X_2$ may be affected by $Z$ and serves as a contributing factor to the noise found in $\hat{Y}$.
Based on the two components, we have the following theorem.

\begin{theorem}~\label{thm:contributing_x2} The instance-dependent causal transition matrix $P(\hat{Y} | do(Y), X)$ is identifiable if we recover the noise predictive factor $X_2$.
\end{theorem}

\noindent\textit{Proof:} Let $G_{\bar{Y}}$ be the graph induced by removing the incoming edges of $Y$. Since $\hat{Y} \bigCI X_1 | Y, X_2$ in $G_{\bar{Y}}$, we have $P(\hat{Y} | do(Y), X) = P(\hat{Y} | do(Y), X_1, X_2) = P(\hat{Y} | do(Y), X_2)$. Let $G_{\underline{Y}}$ be the graph induced by removing the outgoing edges of $Y$. Since $\hat{Y} \bigCI Y | X_2$ in $G_{\underline{Y}}$, we have $P(\hat{Y} | do(Y), X_2) = P(\hat{Y} | Y, X_2)$.
However, note that $Y$ is an unobservable latent variable that we are interested in modeling, and as such we need causal estimand that gives us an unbiased estimation of $Y$ as well.

\begin{theorem}~\label{thm:contributing_x1}
    The effect of $X_1$ on $Y$ can be identified if we recover $X_1$.
\end{theorem}

\noindent\textit{Proof:} Since we have $P(Y | do(X_1)) = \displaystyle\int_{X_2}{P(Y | X_1, X_2) P(X_2)dX_2}$ by using the backdoor criterion and that $Y \bigCI X_2 | X_1$ by d-separation, we have \\
$\displaystyle\int_{X_2}{P(Y | X_1, X_2) P(X_2)dX_2} = \displaystyle\int_{X_2}{P(Y | X_1) P(X_2)dX_2}$ \\
$ = P(Y | X_1) \displaystyle\int_{X_2}{P(X_2)dX_2} = P(Y | X_1)$.


Based on Theorem~\ref{thm:contributing_x1}, it is possible to obtain an unbiased classifier based solely on $X_1$. Consequently, we can recover $X_1$ by decorrelating it from $Z$. 
According to Theorem~\ref{thm:contributing_x2}, the causal transition matrix can be identified if we can identify the contributing factor $X_2 \subseteq X$ to $\hat{Y}$. 
Intuitively, $X_1$ can be omitted, since it is the parent of $Y$ and the $do$ operation effectively eliminates the incoming edge of $Y$. 

Therefore, the focus of this paper is twofold: 1) \textbf{how to identify these contributing factors} and 2) \textbf{ how to model causal relations between variables}.


\subsection{Training Framework for Denoising}
The training framework of our method is illustrated in Figure~\ref{fig:method}. 
In this subsection, we will provide a detailed introduction to the entire procedure.

\noindent\textbf{Modeling the observable contributing factors.}
Initially, we utilize a separation model to separate the noise-resistant component $X_1$ from $X$, which can be represented as $X_1= g_1(X)$, where $g_1(\cdot)$ denotes any neural network.
Subsequently, we infer the clean label $Y$ using a classifier $Y= f(X_1)$.
As mentioned above, $Y$ is unobservable in the causal graph, so it is necessary to sample confidently $Y$ to make it partially observable.
To accomplish this, we employ co-teaching~\cite{han2018co}, as it is a commonly used and straightforward technique.
Two twin models are trained simultaneously and updated via:
\begin{equation}
    \mathcal{L}_{\text{co-teaching}} = \frac{1}{|\mathcal{B}|} \sum_{i \in \mathcal{B}} \min(\ell_{1,i}, \ell_{2,i}),
\end{equation}
where $\mathcal{B}$ denotes the confident examples in a batch with number $|\mathcal{B}|$, and $\ell_{j,i}$ denotes the loss of $i$-th example in $j$-th model.

Since we do not know how the instance $X$ is affected by the latent variable $Z$, we directly model the noise-sensitive component $X_2$ as a representation through $X_2= g_2(X)$, where $g_2(\cdot)$ denotes any type of neural network.

\begin{figure}
\centering
\includegraphics[width=0.8\linewidth]{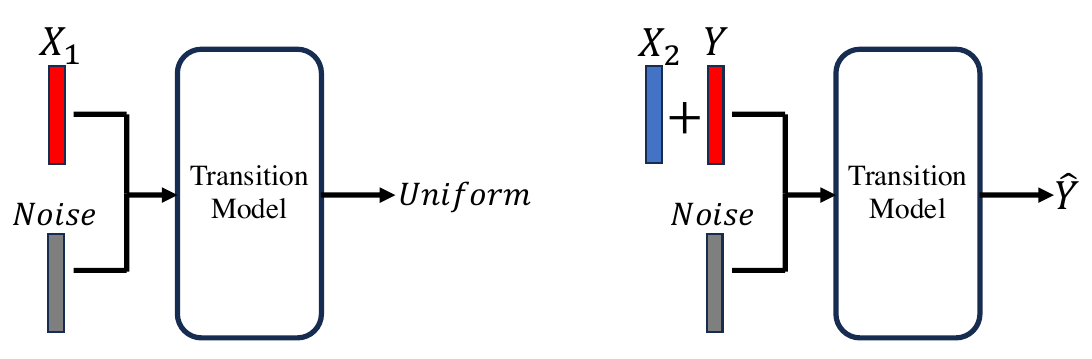}
    \caption{The design of the transition model.}
    \label{fig:transition_model}
\end{figure}

\noindent\textbf{Modeling the causal transition relations.}
To ensure that the variables obey the proposed causal graph, there are two things that must be done.

(i) Decorrelating $X_1$ from $\hat{Y}$.

(ii) Modeling the relationship from $X_2$, $Y$, $Z$ to $\hat{Y}$.

We propose a transition model to achieve both goals simultaneously, as depicted in Figure~\ref{fig:transition_model}, in which the blue block indicates that the gradients do not need to be computed.

Firstly, for decorrelating $X_1$ and $\hat{Y}$, the transition model $f_{tran}$ take $X_1$ and a Gaussian noise as inputs and output a tensor with dimension of $k$, which can be represented as:
\begin{equation}
    \hat{Y}_{X_1}=f_{tran}(X_1, Z_1),~ Z_1\sim\mathcal{N}(0,1),~ \hat{Y}_{X_1}\in \mathbb{R}^k.
\end{equation}
We aim to ensure that the probability of $\hat{Y}$ given $X_1$ is uniform for each class candidate. 
To achieve this, 
we constrain $\hat{Y}_{X_1}$ so that $X_1$ of each instance predicts an all-one vector. This can be represented as the decorrelation loss:
\begin{equation}
    \mathbf{Reg}(X_1)=-\sum_{i=1}^{N} \Tilde{y}_{x_1} \text{log\_softmax}(\mathbf{1}_k),
\end{equation}
where $N$ denotes the number of the examples in a batch, $\Tilde{y}_{x_1}$ denotes the output of the transition model when take $x_1$ as input, and $\mathbf{1}_k$ denotes all-one vector with $k$ dimension.

Secondly, for modeling the causal transition matrix, we take $Y$ and another noise as input, affiliated with the noise-sensitive component $X_2$, which is represented as:
\begin{equation}
    \hat{Y}=f_{tran}(m(Y, X_2), Z_2),~ Z_2\sim\mathcal{N}(0,1),~ \hat{Y}\in \mathbb{R}^k,
\end{equation}
where $m(\cdot,\cdot)$ denotes the merging function of two tensors.
In this paper, we adopt the merging function:
\begin{equation}
\label{eq:gs}
    m(Y,X_2) = gs(Y+\beta * X_2),
\end{equation}
where $\beta$ denotes the scaling factor, and $gs(\cdot)$ denotes the Gumbel-Softmax function.

Since our main focus is on the causal transition matrix $P(\hat{Y}|do(Y),X)$, we block the gradient of $X_2$ to make the relationship from $Y$ to $\hat{Y}$ more precise. 
Therefore, $X_2$ can be considered as the compensation tensor that bridges the gap between $P(\hat{Y}|do(Y),X)$ and $P(\hat{Y}|do(Y))$. The transition model can be optimized using the following equation:
\begin{equation}
        \mathcal{L}_{CE} = -\sum_{i=1}^{N} \Tilde{y}_i \log(\hat{y}_i),
\end{equation}
where $\Tilde{y}$ denotes the predicted noisy label, and $\hat{y}$ denotes the noisy label.

\noindent\textbf{Policy model for getting the noise-sensitive component.}
We still need to optimize $g_2$ to minimize the cross-entropy between the model prediction $\Tilde{Y}$ and the noisy label $\hat{Y}$. However, since we have blocked the gradient of $X_2$ in the transition model, we need to find an alternative approach. Drawing inspiration from reinforcement learning, we use the policy gradient~\cite{sutton2018reinforcement} to obtain the noise-sensitive component $X_2$. 
Specifically, we model $g_2$ using a policy $\pi$ that takes the instance $X$ as input state and outputs a soft action $X_2$. After the soft action is performed in the transition model, a reward $R$ is obtained, representing the outcome of the soft action. We design a reward calculator, which computes the reward for each instance $x_i$ as
$
    R = \frac{1}{1-\Tilde{y}_i \log(\hat{y}_i)}.
$
This reward function ensures that the reward is always greater than zero and increases as the cross-entropy decreases. This naturally promotes the goal of $X_2$ to be achieved. The policy model is trained using the following loss function:
\begin{equation}
    \mathcal{L}_{pg} = -\sum_{i=1}^{N} R \log \pi(x_i).
\end{equation}

Recalling the Gumbel-Softmax function in equation.~\eqref{eq:gs}, it serves two main functions: (1) It acts as an exploration strategy for $\pi$ and prevents the policy from settling into suboptimal results. (2) It introduces more noise into the transition matrix to mimic the effect of $Z$.

\noindent\textbf{Training objectives.}
Our framework can be trained end-to-end, and the overall loss function $\mathcal{L}$ is computed as:
\begin{equation}
    \mathcal{L} = \mathcal{L}_{\text{co-teaching}} + \alpha_1 * \mathcal{L}_{CE} +\alpha_2 * \mathcal{L}_{PG} +\alpha_3 * \mathbf{Reg}(X_1),
    \label{eq:total_loss}
\end{equation}
where $\alpha_1$, $\alpha_2$ and $\alpha_3$ denotes the scale factors.

\subsection{Analysis of the Framework}

\textbf{What our framework does?} Theories \ref{thm:contributing_x2} and \ref{thm:contributing_x1} assert that under the proposed causal graph, the causal transition matrix is identifiable and enables us to infer the clean label $Y$ during the inference process. Our framework establishes the causal relationship through Equation~\eqref{eq:total_loss}, which comprises four terms of loss functions. The co-teaching loss $\mathcal{L}_{\text{co-teaching}}$ treats the confident noisy label $\hat{Y}$ as the true label $Y$ and guides the training of the classifiers within the twin networks. Concurrently, the policy gradient loss $\mathcal{L}_{PG}$ is instrumental in separating the noise-sensitive component $X_2$ from the input instance $X$. Additionally, the regularization loss $\mathbf{Reg}(X_1)$ acts as a penalty to reduce the correlation between $X_1$ and $\hat{Y}$, ensuring that the noise-resistant component $X_1$ remains insensitive to noise.
Our proposed causal graph accommodates the linkage between the two variables $X_1$ and $X_2$, obviating the need for additional computations to disentangle them. Once the variables are distinguished, the causal transition relationship is explicitly established through the transition model, which is refined by optimizing the cross-entropy loss $\mathcal{L}_{CE}$.

\noindent\textbf{Additional computational costs.} During the training process, we construct a twin network architecture, which requires twice the computational resources compared to the original classifier. However, for inference, only the separation model and the classifier are employed. Since the separation model can be implemented as a relatively simple layer, the increase in computational costs remains minimal.
Furthermore, our method does not adopt generative models, unlike previous work~\cite{yao2021instance,bae2022noisy,cheng2020learning}, which not only saves computational costs but also eliminates potential errors.

\noindent\textbf{Relation to mainstream assumptions.} Our method is adept at addressing both instance-independent and instance-dependent label noise. By focusing solely on the right part of the causal graph and disregarding $X_2$, our approach aligns with the \textbf{instance-independent assumption} that 
$P(\hat{Y}|Y, X) = P(\hat{Y}|Y)$. This equivalence arises due to the decorrelation between $X_1$ and $\hat{Y}$.
Furthermore, our method does not rely on the \textbf{assumption of independence in the posterior probability} 
$P(Y|\hat{Y}, X) = P(Y|\hat{Y})$, as seen in related work~\cite{yao2021instance,bae2022noisy}.
Although most related work depends on specific assumptions, our approach is based on two relatively mild and broadly applicable assumptions: the separability of
$X$ and the partial observability of $Y$. These assumptions are generally reasonable and pragmatic for many real-world applications.

\noindent\textbf{Flexibility of the causal graph.} The proposed causal graph can also be adapted for semi-supervised methods~\cite{wang2022debiased,xiao2023promix,li2020dividemix}, which initially select clean samples and then use the remaining noisy samples to enhance performance.
Further details on the implementation and experimental results will be provided in the Appendix.

\section{Experiment}
\label{sec:exp}
We carry out experiments on both \textbf{synthetic} and \textbf{real-world datasets}, encompassing various types of label noise.
We place particular emphasis on instance-dependent noise, as it represents the most challenging and significant aspect of our research.

\subsection{Experiment Setup}

\textbf{Datasets.}
(i) We first perform experiments on the manually corrupted version of four \textbf{synthetic datasets},~\ie, FashionMNIST~\cite{xiao2017fashion}, SVHN~\cite{yuval2011reading}, CIFAR10, CIFAR100~\cite{krizhevsky2009learning}.
The experiments are conducted with three different types of artificial label noise, with a focus on instance-dependent noise in this paper.
(a) \textbf{Symmetric Noise~(SYM)}. The label of each instance is uniformly flipped to one of the other classes~\cite{patrini2017making,han2018co,xia2020robust}.
(b) \textbf{Asymmetric Noise~(AYM)}. The label of each instance is flipped to a set of semantically similar classes~\cite{tanaka2018joint,han2018co,xia2020robust}.
(c) \textbf{Instance Dependent Noise~(IDN)}. The label of each instance is determined by the probability that it is mislabeled, and this probability is computed based on the corresponding feature of the data instance~\cite{berthon2021confidence,yao2021instance,bae2022noisy}.

(ii) Meanwhile, we perform experiments on two \textbf{real-world} noisy datasets: Food101~\cite{bossard2014food} and Clothing1M~\cite{xiao2015learning}.

\noindent\textbf{Evaluation metric.} \textbf{The test sets for the various datasets remain clean}, ensuring that the \textbf{test accuracy} can effectively reflect the superior performance of the denoising methods we evaluate.
For the synthetic datasets, we report the mean performance across 5 random seeds.
For the real-world dataset, we report the best results for the last 10 epochs.

\noindent\textbf{Baselines.}
We choose popular denoising methods as baselines: Co-teaching~\cite{han2018co}, Joint~\cite{tanaka2018joint}, JoCoR~\cite{wei2020combating}, CORES2~\cite{cheng2020learning}, SCE~\cite{wang2019symmetric}, LS~\cite{lukasik2020does}, REL~\cite{xia2020robust}, Forward~\cite{patrini2017making}, DualT~\cite{yao2020dual}, TVR~\cite{zhang2021learning}, MentorNet~\cite{jiang2018mentornet}, Mixup~\cite{zhang2018mixup}, Reweight~\cite{liu2015classification}, T-Revision~\cite{xia2019anchor}, 
BLTM-V~\cite{yang2022estimating},
CausalNL~\cite{yao2021instance}.
Among them, CausalNL is closest to our work, since it is also a causal-based method.

\noindent\textbf{Ablation study.}
We also perform ablations on all scenarios by removing the policy model $X_2$ from the framework.
This indicates that our model is trained with a transition relationship that follows the instance-independent assumption $P(\hat{Y}|Y,X)=P(\hat{Y}|Y)$.
We represent this as ``w.o/pg'' in this section.

\noindent\textbf{Implementation details.}
We adopt the residual network models~\cite{he2016deep} as classifiers in our experiment, and the depth of the neural network increases with the difficulty of the dataset.
The experimental parameter settings are also slightly varied towards different datasets, please refer to the Appendix for more details.

\setlength{\tabcolsep}{6pt}
\begin{table}[t]
\caption{Results on FashionMNIST with symmetric, asymmetric, and instance-dependent label noise.}
\vspace{-1em}
\fontsize{8}{13}\selectfont
\centering
\begin{tabular}{l|cc|cc|cc}
\hline
\multirow{2}{*}{Method}          
& \multicolumn{2}{c|}{SYM}    & \multicolumn{2}{c|}{ASYM}   & \multicolumn{2}{c}{IDN}    \\ 

& 20\% &80\%  &20\% &40\%   &20\% &40\%  \\ 
\hline
CE  &74.0  &27.0 &81.0 &77.3  &68.4  &52.1   \\
Early Stop &83.6 &49.5 &84.1 &76.6 &79.5 &55.4   \\
Co-teaching &82.5 &64.2 &88.2 &73.6 &81.8 &75.4  \\
Joint &82.0 &6.0 &82.1 &82.3 &82.7 &82.4  \\
JoCoR &86.0 &27.6 &88.9 &79.4 &86.3 &83.2    \\
CORES2 & 74.6 & 8.9  &77.6 & 74.3 &80.0 &58.1  \\
SCE & 74.0 & 27.0  &82.0 & 77.4 & 68.3 & 52.0   \\
LS & 73.9 & 27.8  &81.5 & 77.0 & 69.0 & 52.5   \\
REL &84.6 &70.1 &82.8 &76.2 &84.6 &75.5   \\
Forward   &77.4 &24.3 &88.3 &79.2 &75.2 &56.9   \\
DualT  &84.5 &10.0 &86.9 &83.1 &85.1 &68.5   \\
TVR  &72.6 &24.9 &80.6 &76.4 &66.3 &51.7   \\
CausalNL  &84.0 &51.5 &88.8 &87.4 &90.8 &90.0   \\ \hline
Ours w.o/ pg  &\textbf{92.1} &\textbf{75.8} &89.6 &82.4 &\textbf{91.3 }&90.3   \\
Ours &\textbf{92.1} &71.5 &\textbf{91.4} &\textbf{88.7} &91.0 &\textbf{90.4 }  \\
\hline
\end{tabular}
\label{tab:fmnist}
\vspace{-1em}
\end{table}

\subsection{Results on Symmetric and Asymmetric Noise}
We begin our experiments with the relatively simple dataset FashionMNIST, considering both symmetric and asymmetric noise. The results for different noise rates are shown in Table~\ref{tab:fmnist}.
For symmetric noise, we conduct experiments with noise rates of 20\% and 80\%. 
For asymmetric noise, we conduct experiments with noise rates of 20\% and 40\%.
In both scenarios, our ablation study achieves the best performance.
The reason behind this is that the noisy labels are randomly assigned without considering the instance, making them almost instance-independent, where $P(\hat{Y}|Y,X) = P(\hat{Y}|Y)$.
The policy model can be considered as $P(\hat{Y}|Y+\mathcal{R})$, where a real number $\mathcal{R}$ is added to $Y$, thus affecting the model's performance.

\subsection{Results on Instance-dependent Noise}

\noindent\textbf{Comparison with baselines.} Instance-dependent noise is crucial for evaluating the accuracy of the estimated transition matrix, as it is challenging to satisfy the instance-independent assumption.
We conducted our study on four datasets: FashionMNIST (Table~\ref{tab:fmnist}), SVHN (Table~\ref{tab:svnh}), CIFAR10, and CIFAR100 (Table~\ref{tab:cifar}), with increasing levels of difficulty.
As the results demonstrate, our method achieves state-of-the-art performance in almost all scenarios, particularly when confronted with high noise rates.
Even the ablation study of our method yields satisfactory results. This can be attributed to the separation of the $X_1$ component from $X$, which allows for explicit decorrelation between $X_1$ and the noisy label $\hat{Y}$, resulting in a more reliable classifier.


\begin{table}[t]
\caption{Results on SVNH with instance-dependent noise.}
\vspace{-1em}
\fontsize{7}{13}\selectfont
\setlength{\tabcolsep}{1.8pt}
\centering
\begin{tabular}{l|ccccc}
\hline
\multirow{2}{*}{Method}            & \multicolumn{5}{c}{IDN } \\ 
& 20\% &30\%  &40\% &45\%   &50\%  \\ 
\hline
CE &91.51$\pm$0.45  & 91.21$\pm0.43$ &87.87$\pm1.12$ &67.15$\pm1.65$  & 51.01$\pm3.62$   \\
Co-teaching &93.93$\pm$0.31  &92.06$\pm$0.31 &91.93$\pm$0.81 &89.33$\pm$0.71 &67.62$\pm$1.99 \\
Decoupling &90.02$\pm$0.25  &91.59$\pm$0.25 &88.27$\pm$0.42 &84.57$\pm$0.89 &65.14$\pm$2.79  \\
MentorNet & 94.08$\pm$0.12 & 92.73$\pm$0.37 &90.41$\pm$0.49 &87.45$\pm$0.75 &61.23$\pm$2.82 \\
Mixup & 89.73$\pm$0.37 & 90.02$\pm$0.35  &85.47$\pm$0.63 &82.41$\pm$0.62 &68.95$\pm$2.58   \\
Forward &91.89$\pm$0.31 &91.59$\pm$0.23 &89.33$\pm$0.53 &80.15$\pm$1.91 &62.53$\pm$3.35   \\
Reweight &92.44$\pm$0.34 &92.32$\pm$0.51 &91.31$\pm$0.67 &85.93$\pm$0.84 &64.13$\pm$3.75  \\
T-Revision  &93.14$\pm$0.53 &93.51$\pm$0.74  &92.65$\pm$0.76 &88.54$\pm$1.58 &64.51$\pm$3.42  \\
BLTM-V  &\textbf{95.12$\pm$0.40} & \textbf{94.69$\pm$0.24} & 88.13$\pm$3.23  &80.43$\pm$4.12  &78.71$\pm$4.37\\
CausalNL  &94.06$\pm$0.23 &93.86$\pm$0.65& 93.82$\pm$0.64 &93.19$\pm$0.93 &85.41$\pm$2.95  \\ \hline
Ours w.o/ pg &93.86$\pm$0.17 &93.82$\pm$0.18 &93.50$\pm$0.25 &93.19$\pm$1.25 &\textbf{92.91$\pm$1.54}   \\
Ours  &94.13$\pm$0.08 &93.97$\pm$0.11 &\textbf{93.94$\pm$0.16} &\textbf{93.33$\pm$1.12} &92.57$\pm$1.56   \\
\hline
\end{tabular}
\label{tab:svnh}
\end{table}

\begin{table*}[!ht]
\centering
\caption{Results on CIFAR dataset.}
\vspace{-1em}
\fontsize{8}{12}\selectfont
\setlength{\tabcolsep}{1.8pt}

\begin{tabular*}{0.98\textwidth}{@{\extracolsep{\fill}} l|ccccc|ccccc @{}}
\hline
\multirow{2}{*}{Method}          
& \multicolumn{5}{c|}{CIFAR10-IDN}    & \multicolumn{5}{c}{CIFAR100-IDN}     \\ 
& 20\% &30\%  &40\% &45\%  &50\%   &20\% &30\%  &40\% &45\%  &50\% \\ \hline
CE &75.81$\pm$0.26 & 69.15$\pm$0.65 &62.45$\pm$0.86 &51.72$\pm$1.34  & 39.42$\pm$2.52 &30.42$\pm$0.44 &24.15$\pm$0.78 &21.45$\pm$0.70 &15.23$\pm$1.32 &14.42$\pm$2.21   \\
Co-teaching &80.96$\pm$0.31  &78.56$\pm$0.61 &73.41$\pm$0.78 &71.60$\pm$0.79 &45.92$\pm$2.21  &37.96$\pm$0.53 &33.43$\pm$0.74 &28.04$\pm$1.43 &25.60$\pm$0.93 &23.97$\pm$1.91\\
Decoupling &78.71$\pm$0.15  &75.17$\pm$0.58 &61.73$\pm$0.34 &58.61$\pm$1.73 &50.43$\pm$2.19  &36.53$\pm$0.49 &30.93$\pm$0.88 &27.85$\pm$0.91 &23.81$\pm$1.31 &19.59$\pm$2.12  \\
MentorNet & 81.03$\pm$0.24 & 77.22$\pm$0.47 &71.83$\pm$0.49 &66.18$\pm$0.64 &47.89$\pm$2.03 &38.91$\pm$0.54 &34.23$\pm$0.73 &31.89$\pm$1.19 &27.53$\pm$1.23 &24.15$\pm$2.31 \\
Mixup & 73.17$\pm$0.34 & 70.02$\pm$0.31  &61.56$\pm$0.71 & 56.45$\pm$0.67 & 48.95$\pm$2.58 &32.92$\pm$0.76 &29.76$\pm$0.87 &25.92$\pm$1.26 &23.13$\pm$2.15 &21.31$\pm$1.32   \\
Forward &76.64$\pm$0.26 &69.75$\pm$0.56 &60.21$\pm$0.75 &48.81$\pm$2.59 &46.27$\pm$1.30 &36.38$\pm$0.92 &33.17$\pm$0.73 &26.75$\pm$0.93 &21.93$\pm$1.29 &19.27$\pm$2.11   \\
Reweight &76.23$\pm$0.25 &70.12$\pm$0.72 &62.58$\pm$0.46 &51.54$\pm$0.92 &45.46$\pm$2.56  &36.73$\pm$0.72 &31.91$\pm$0.91 &28.39$\pm$1.46 &24.12$\pm$1.41 &20.23$\pm$1.23  \\
T-Revision  &76.15$\pm$0.37 &70.36 $\pm$0.54&64.09$\pm$0.37 &54.42$\pm$1.01 &49.02$\pm$2.13  &37.24$\pm$0.85 &36.54$\pm$0.79 &27.23$\pm$1.13 &25.53$\pm$1.94 &22.54$\pm$1.95  \\

BLTM-V\footnote{We apologize that were unable to find suitable hyperparameters to achieve a desirable outcome on CIFAR100 with BLTM-V.} & 80.37$\pm$1.98 &78.82$\pm$1.07 &72.93$\pm$4.00 &64.83$\pm$4.65& 60.33$\pm$5.29& - &- &- &-& -\\
CausalNL  &81.47$\pm$0.32 &80.38$\pm$0.44 &77.53$\pm$0.45 &78.60$\pm$1.06 &77.39$\pm$1.24  &41.47$\pm$0.43 &40.98$\pm$0.62 &34.02$\pm$0.95 &33.34$\pm$1.13 &32.13$\pm$2.23  \\ \hline
Ours w.o/ pg &82.57$\pm$0.33 &81.24$\pm$0.36 &79.36$\pm$0.81 &78.43$\pm$0.51 &75.59$\pm$2.07  &45.50$\pm$0.99 &\textbf{44.67$\pm$0.60} &38.44$\pm$1.40 &34.88$\pm$2.53 &33.05$\pm$1.47   \\
Ours  &\textbf{82.94$\pm$0.29} &\textbf{82.15$\pm$0.25} &\textbf{81.04$\pm$0.23} &\textbf{80.24$\pm$0.39} &\textbf{78.37$\pm$0.93} &\textbf{46.37$\pm$0.46} &43.34$\pm$0.39 &\textbf{39.61$\pm$1.04} &\textbf{37.04$\pm$1.83} &\textbf{34.44$\pm$1.86}   \\
\hline
\end{tabular*}
\label{tab:cifar}
\end{table*}

\begin{table}[h]
\caption{Results on real-world dataset.}
\vspace{-1em}
\fontsize{9}{12}\selectfont
\centering
\begin{tabular}{l|cc}
\hline
\multirow{2}{*}{Method}         
&  \multicolumn{2}{c}{Dataset}     \\ 
& \multicolumn{1}{c}{Food101}    & \multicolumn{1}{c}{Clothing1M}     \\ 
\hline
CE  &78.37 & 68.88   \\
Early Stop &73.22 &67.07   \\
Co-teaching & 78.35 &60.15 \\
SCE   & 75.23 & 67.77   \\
REL &78.96 &62.53    \\
Forward  &83.76 &69.91     \\
DualT  &57.46 &70.18     \\
TVR  &77.37 &69.44  \\
CausalNL  &85.64 &68.90   \\ \hline
Ours w.o/ pg &85.52 &70.48   \\
Ours  &\textbf{85.86} &\textbf{72.25 }  \\
\hline
\end{tabular}
\label{tab:real}
    \vspace{-2em}

\end{table}

\begin{figure}[h]
    \centering
    \begin{subfigure}{0.48\linewidth}
     \centering
        \includegraphics[width=\linewidth]{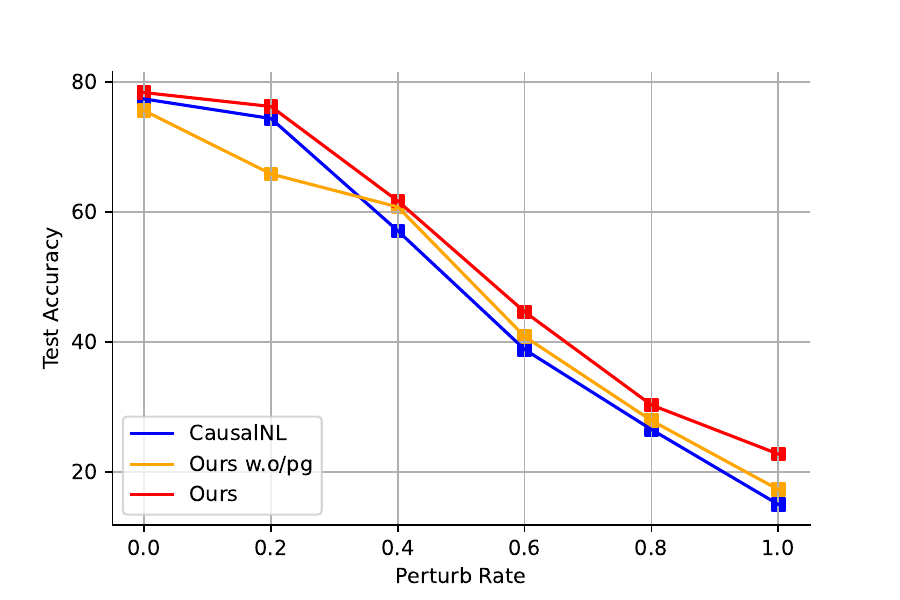}
        \caption{Perturbing the training instances considering different intensities  with a 50\% instance-dependent noise.}
    \end{subfigure}
    \centering
    \begin{subfigure}{0.48\linewidth}
     \centering
        \includegraphics[width=\linewidth]{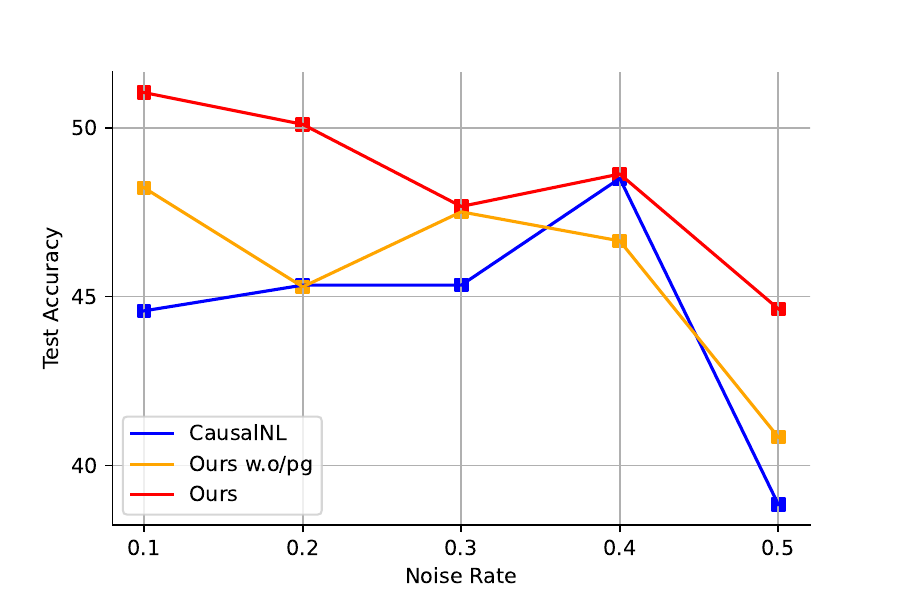}
        \caption{Considering different levels of instance-dependent noise when perturbing the instances with an intensity of $0.6$.  }
    \end{subfigure}
    \centering
    \caption{Model performance on the CIFAR10 Dataset.}
    \label{fig:noise_demo}
       \vspace{-1em}
\end{figure}

\noindent\textbf{How the latent variable $Z$ influence the performance.}
In real-world scenarios, the latent variable $Z$ influences both the instance $X$ and the noisy label $\hat{Y}$. In this subsection, we introduce additional noise to perturb the instance $X$, reflecting its influence along the path $Z \rightarrow X$ in the causal graph. Meanwhile, the noise rate represents the influence of $Z$ along the path $Z \rightarrow \hat{Y}$. To investigate how the latent variable $Z$ influences performance, we control the influence of $Z$ on one path while adjusting the other.

\noindent\textbf{Construction of the perturbed instance.} 
We introduce noise to each instance $X$ in the training data by generating a noise vector $Z_{\text{noise}}$ from a uniform distribution, $Z_{\text{noise}} \sim U([0, 1])$, where $Z_{\text{noise}} \in \mathbb{R}^D$, and $D$ denotes the dimensionality of instance $X$.
The perturbed instance $X_{\text{per}}$ is generated by:
$
X_{\text{per}} = X + \gamma \cdot Z_{\text{noise}},
$
where $\gamma$ is the scale factor that determines the intensity of the perturbation.

\noindent\textbf{How the performance decreases.} Figure~\ref{fig:noise_demo} (a) shows the performance curve as we gradually increase the intensity of the perturbation $\gamma$ from 0 to 1 on the CIFAR10 dataset with a 50\% instance-dependent noisy label rate.
As the perturbation increases, the performance of all methods decreases, but our method consistently exhibits the best performance.
Figure~\ref{fig:noise_demo} (b) illustrates the performance curve as we increase the noisy label rate from 0.1 to 0.5 while keeping $\gamma$ fixed at 0.6.
Compared to CausalNL and the ablation study, our method demonstrates not only higher effectiveness but also greater stability.
In particular, the performance of both baselines experiences a significant drop when the label noise rate increases from 0.4 to 0.5, while our method is relatively unaffected.
This can be attributed to our modeling of the noise-sensitive component $X_2$, which helps bridge the gap between $P(\hat{Y}|Y)$ and $P(\hat{Y}|Y,X)$ in certain contexts.

\subsection{Results on Real-world Noisy Dataset}
We also performed experiments on two popular noisy real-world label datasets, Food101 and Clothing1M, to evaluate the superiority of our method.
Table~\ref{tab:real} presents the test accuracy on the clean test set.
Our method outperforms the baselines, demonstrating its effectiveness in real-world settings.
Even in the ablation study, our method without the policy model still achieves competitive performance compared to the baseline.
This is primarily due to our explicit separation of the $X_1$ component from the original instance $X$, which is not influenced by the latent variable $X$ and directly contributes to the learning of the classifier.
The ablation results also indicate that the label noise in Clothing1M is more likely to be instance-dependent compared to Food101.

\section{Conclusion}
\label{sec:conclusion}
In this paper, our objective is to learn an accurate transition relationship in label-noise learning to obtain a better classifier. 
To gain a deeper understanding of the label noise generation procedure, we approa the problem from a causal viewpoint and propose a novel causal structure for learning with label-noise. 
Within this causal structure, we introduce the concept of a ``causal transition matrix'' $P(\hat{Y}|do(Y),X)$, which can approximate the original transition matrix $P(\hat{Y}|Y,X)$ without relying on the instance-independent assumption. 
To accomplish this,  we developed a framework that enables us to estimate the causal transition matrix with a theoretical guarantee of identifiability. Experimental results on both synthetic and real-world datasets validated the superior performance of our method.
\section*{Acknowledgements}
This work was supported by the National Natural Science Foundation of China (62376243, 62441605, 62037001), the Starry Night Science Fund at Shanghai Institute for Advanced Study (Zhejiang University), and Ant Group Postdoctoral Programme. 
Long Chen was supported by RGC Early Career Scheme (26208924), National Natural Science Foundation of China Young Scientist Fund (NSFC24EG42), and HKUST Sports Science and Technology Research Grant (SSTRG24EG04).


%
%
\bibliography{aaai25}

\clearpage
\appendix

\section{Comparison with Causal-based Methods}
\begin{figure}[h]
     \centering
        \includegraphics[width=0.85\linewidth]{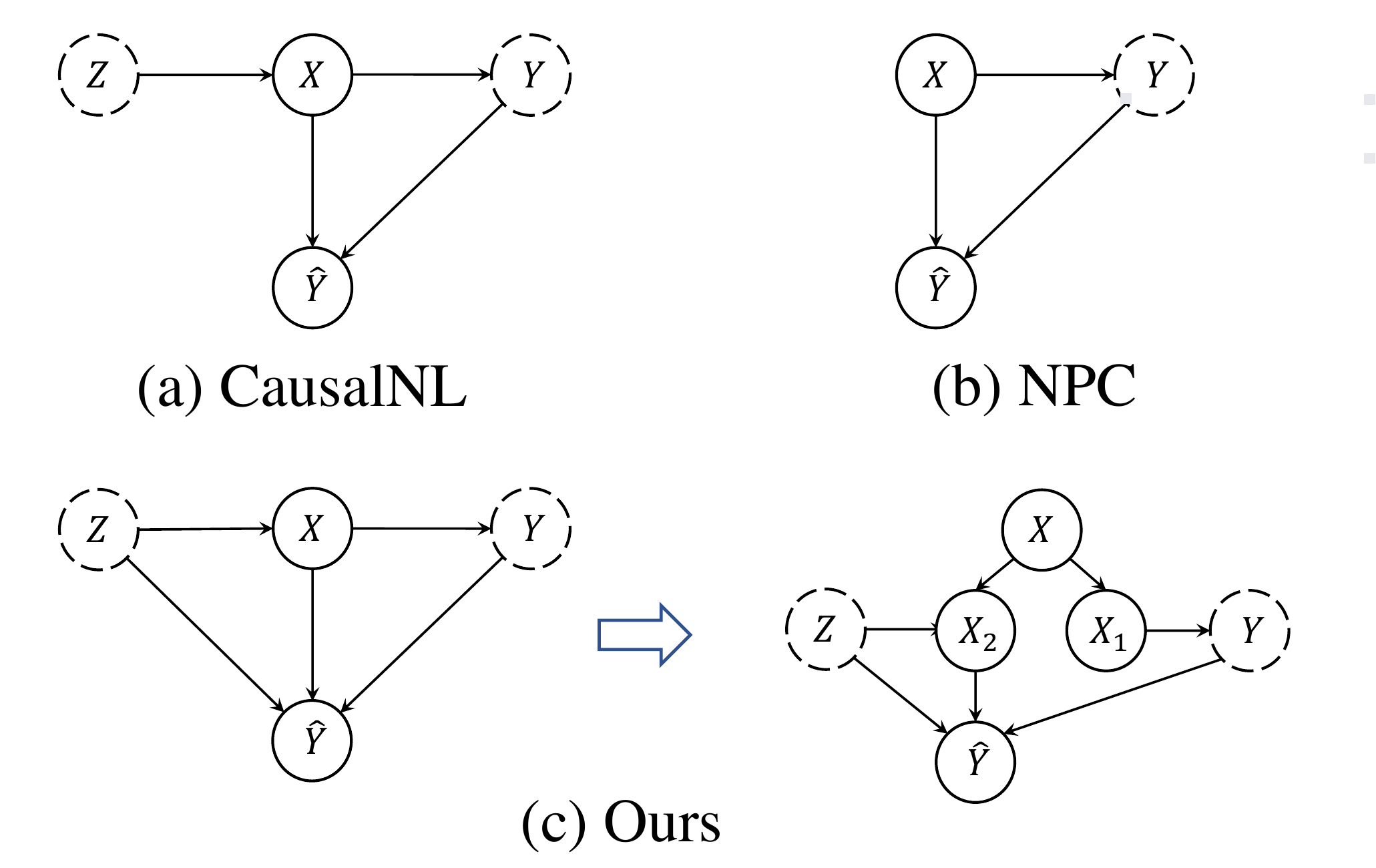}
  \caption{Causal graphs for data generation process.}
    \label{fig:cgs}
\end{figure}

The causal perspective is instrumental in dissecting the data generation process, particularly in the presence of noise. In addition to our approach, two prominent causal-based methods are CausalNL~\cite{yao2021instance} and NPC~\cite{bae2022noisy}, whose proposed causal graphs are depicted alongside ours in Figures~\ref{fig:cgs}. Upon examining these graphs, it becomes evident that CausalNL overlooks the direct causal influence of $Z$ on $\hat{Y}$, whereas NPC intentionally omits $Z$. However, both methods presuppose instance independence within the\textbf{ posterior probability}, $P(Y|\hat{Y},X) = P(Y|\hat{Y})$, and rely on generative models such as VAEs (Variational Autoencoders)~\cite{kingma2013auto} to establish relationships among the variables.

In marked contrast, our methodology dispenses with the need for generative models and discards the assumption of instance-independent posterior probability. Operating within this framework, we introduce the concept of ``causal transition matrix'' $P(\hat{Y}|do(Y),X)$, which seeks to accurately estimate the authentic transition matrix $P(\hat{Y}|Y,X)$ and comes with theoretical guarantee, thus enhancing the effectiveness of learning under noise conditions.
We \textbf{do not include a comparison with NPC} in our main paper, as \textbf{it is a post-hoc method} that is applied directly to any classifiers, including our own.

\begin{figure*}[h]
    \centering
    \begin{subfigure}{0.325\linewidth}
     \centering
        \includegraphics[width=\linewidth]{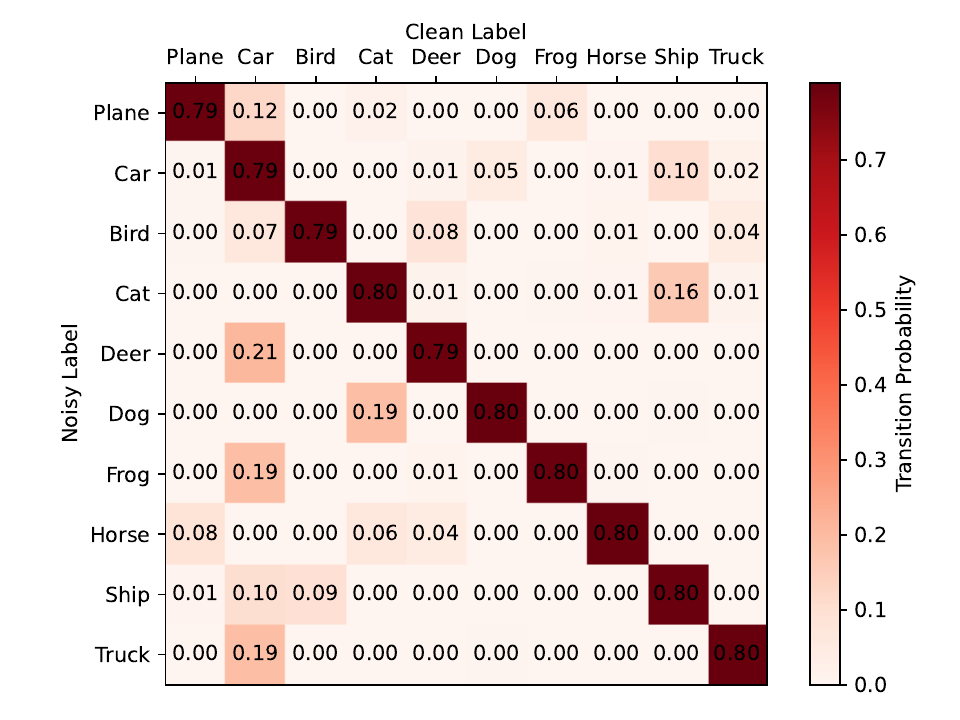}
        \caption{}
    \end{subfigure}
    \centering
    \begin{subfigure}{0.325\linewidth}
     \centering
        \includegraphics[width=\linewidth]{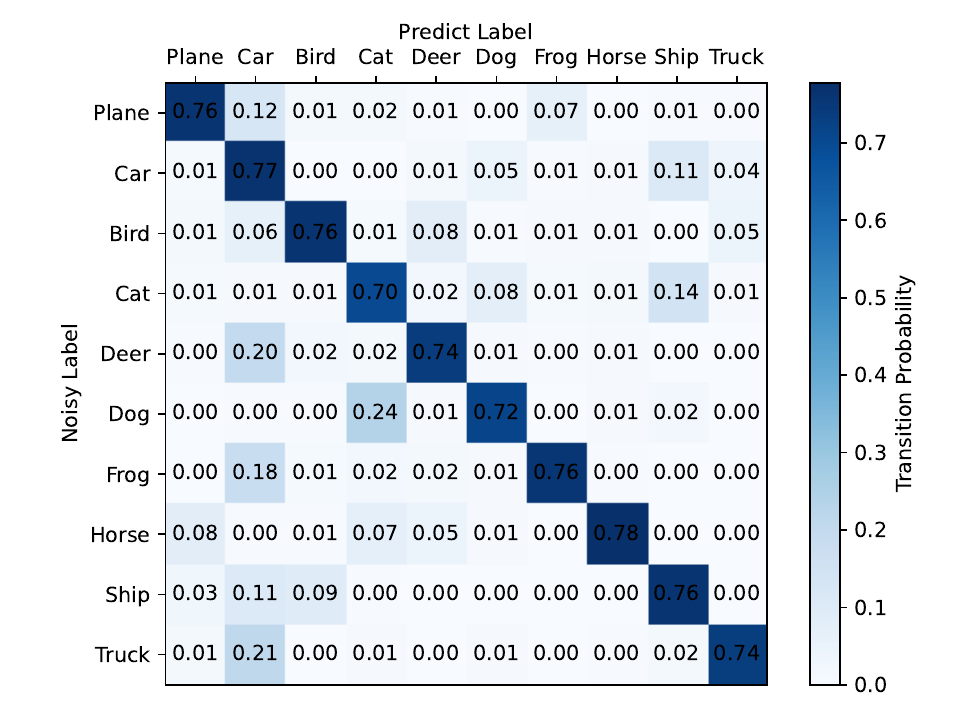}
        \caption{}
    \end{subfigure}
    \centering
    \begin{subfigure}{0.325\linewidth}
     \centering
        \includegraphics[width=\linewidth]{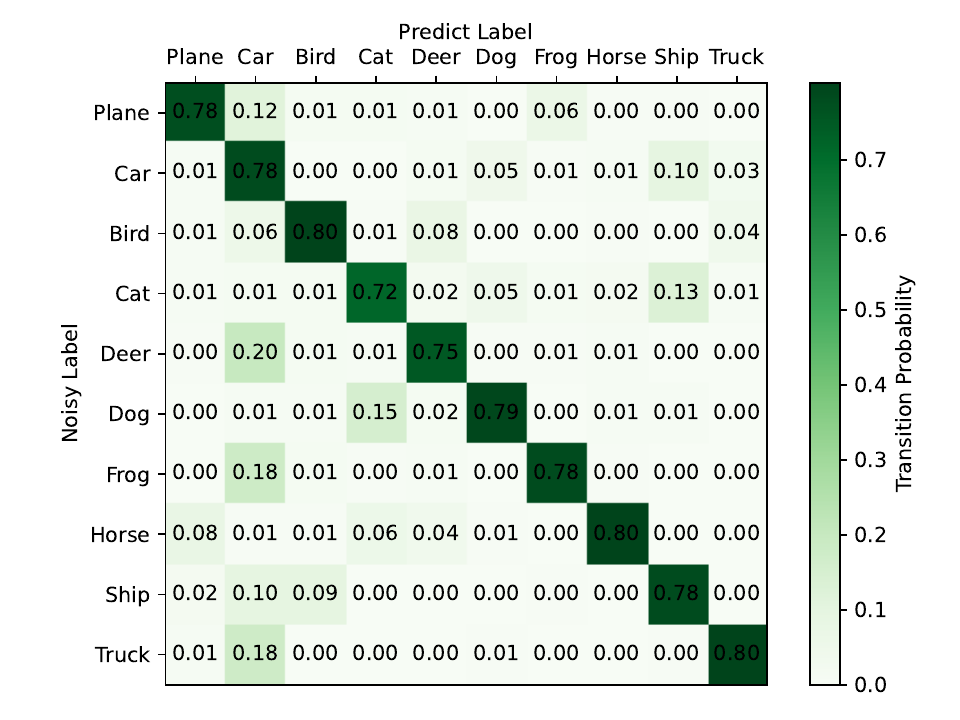}
        \caption{}
    \end{subfigure}
    \caption{(a) The ground truth transition matrix. (b) The transition matrix of our method. (c) The transition matrix of our ablation study. }
    \label{fig:transition_matrix}
\end{figure*}

\section{Viewing the Transition Relationship} To demonstrate the effectiveness of our method more intuitively, we present the estimated transition matrix on CIFAR10 with a 20\% instance-dependent noisy rate in Figure~\ref{fig:transition_matrix}.
As shown in the figure, most of the clean instances are correctly classified, indicating that our method avoids overfitting to the noisy labels.
Additionally, our method is capable of correcting for some mislabeled instances.
The ablation study also exhibits promising performance in inferring clean instances, which highlights the usefulness of separating $X_1$ from $X$.
However, the transition matrix also reveals some failure cases, such as the inability to correct instances labeled as ``Dog'' to ``Cat'', which may be due to their semantic similarity.

\begin{figure}[h]
     \centering
        \includegraphics[width=0.8\linewidth]{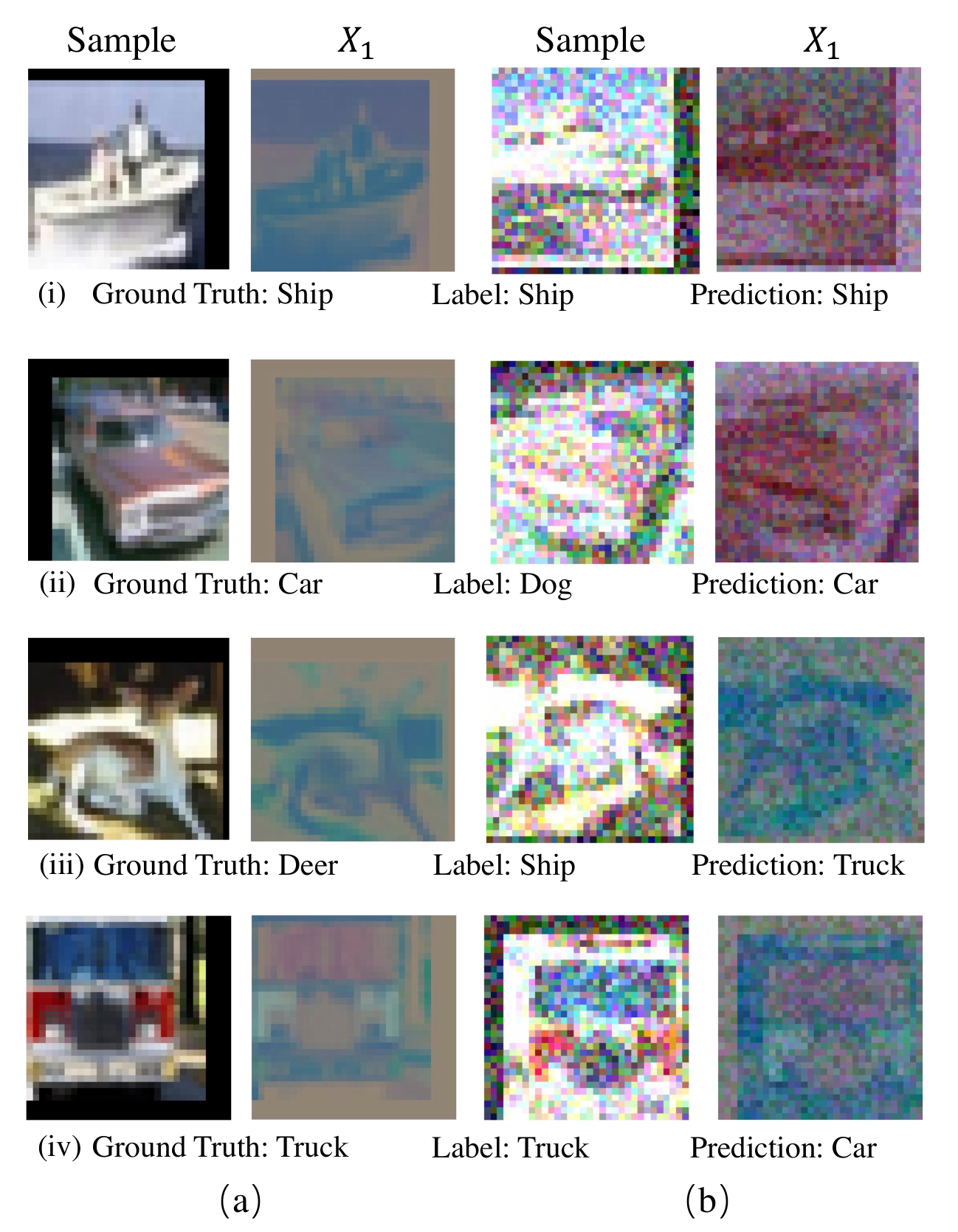}
        \vspace{-1em}
  \caption{Samples and feature map of $X_1$ in (a) CIFAR10 with 30\% instance-dependent label noise (b) CIFAR10 with 30\% instance-dependent label noise and perturbation of 0.6 intensity.}
    \label{fig:x1}
\end{figure}

\section{Viewing What $X_1$ Learned}
Our method is based on two variables, $X_1$ and $X_2$, in the causal graph, which represent the noise-resistant and noise-sensitive component of $X$, respectively.
In our proposed framework, $X_2$ serves as an abstract representation, while $X_1$ is modeled using a single $1*1$ convolution layer without an activation function.
To investigate what $X_1$ has learned, we visualize the feature map of $X_1$ and display representative examples in Figure~\ref{fig:x1}.
From the figure, we observe that $X_1$ preserves the contour information of the image while neglecting some color information in this dataset.
Given that instance-dependent label noise is generated based on the features of each instance, it is reasonable for the model to learn a feature map that relates $X_1$ to the clean label $Y$, but not to the noisy label.
Moreover, discarding color information can also be beneficial in improving the model's performance when faced with perturbations on the instance in certain contexts.
The third and fourth columns illustrate the failure cases of our method.
In the third column, the ground truth label of the image is ``Deer,'' but it is labeled as ``Ship.''
Our method fails to correct the label and predicts the image as ``Truck,'' even without any instance perturbation.
Given that the transition probability for this case is close to zero, the failure may be attributed to the model itself rather than the denoise technology.
In the fourth column, the ground truth label and the image training label are both ``Truck,'' but the model predicts it as ``Ship.''
Upon examining the ground truth transition matrix and the estimated transition matrix, we observe a relatively high probability of transitioning from ``Truck'' to ``Car.''
As a result, our model may incorrectly predict the clean instance with a probability of 1\%.

\section{Training Details}
All experiments in this paper were carried out using an NVIDIA V100 GPU with 32GB of memory.
The depth of the residual network models~\cite{he2016deep} increases with the difficulty of the dataset. For FASHIONMNIST, we use ResNet18.
For SVHN and CIFAR10, we used ResNet34.
For CIFAR100, we used ResNet50 without pretrained checkpoints.
For Food101 and Clothing1M, we use ResNet50 with pretrained checkpoints.

For synthetic datasets, the learning rate is set to $1e^{-3}$, and for real-world datasets, the learning rate is set to $1e^{-4}$.
For synthetic datasets and Food101, we train for 150 epochs, and for the Clothing dataset, we train for 20 epochs.
We set $\alpha_1=\alpha_3=0.1$ for all scenarios.
The value of $\alpha_2$ depends on the difficulty of the scenarios.
For datasets without additional perturbations on the instances, we set $\alpha_2=0.1$.
For the datasets with perturbations, we set $\alpha_2=0.01$.
In addition, we set $\beta=0.2$ in all experiments. Adam optimizer~\cite{kingma2014adam} is used in all experiments in this paper.

\begin{table*}[h]
\caption{Experimental settings.}
\centering
\begin{tabular}{l|cccccc}
\hline
Dataset &FMNIST &CIFAR10 &CIFAR100 & SVNH & Food101 & Clothing1M  \\
\hline
Backbone  &ResNet18 & ResNet34 & ResNet50& ResNet34& ResNet50*& ResNet50*\\
Separation Model  &1*1 Conv &1*1 Conv&1*1 Conv&1*1 Conv& 1*1 Conv&1*1 Conv\\
Policy Model  &4 Conv & 4 Conv & 4 Conv& 4 Conv&5  Conv& 5 Conv\\
Learning Rate  & $1e^{-3}$ &  $1e^{-3}$ & $1e^{-3}$ & $1e^{-3}$ & $1e^{-4}$ & $1e^{-4}$    \\
Epochs &150 & 150 &150 & 150 &20 & 20  \\
$\alpha_1$ &0.1 & 0.1 &0.1 & 0.1 &0.1 & 0.1  \\
$\alpha_2$\footnote{We set $\alpha_2=0.01$ when the instance $X$ is perturbed.} &0.1 & 0.1 &0.1 & 0.1 &0.1 & 0.1  \\
$\alpha_3$ &0.1 & 0.1 &0.1 & 0.1 &0.1 & 0.1  \\
$\beta$ &0.2 & 0.2 &0.2 & 0.2 &0.2 & 0.2  \\

\hline
\end{tabular}
\label{tab:hyper}
\end{table*}

For each dataset, we consistently employ the same learning rate and backbone across all methods. Our training hyperparameters \footnote{The superscript $^*$ indicates that we utilize the pretrained backbone.} are detailed in Table~\ref{tab:hyper}. In the separation model for deriving $X_1$, we utilize a singular 1×1 convolutional layer. For the policy model that computes $X_2$, we employ four convolutional layers with channel dimensions [32, 64, 128, 256] - defined as ``4 Conv'' in the table - for FMNIST~\cite{xiao2017fashion}, CIFAR~\cite{krizhevsky2009learning}, and SVHN~\cite{yuval2011reading} datasets. 
Alternatively, for Food101~\cite{bossard2014food}, and Clothing1M~\cite{xiao2015learning} datasets, we apply five convolutional layers with output channels [32, 64, 128, 256, 512] - called ``5 Conv'' in the table. We maintain a consistent kernel size of 3 and stride of 2 across all configurations.

\setlength{\tabcolsep}{6pt}
\begin{table}[t]
\caption{Results on CIFAR-N dataset}
\fontsize{9.4}{13}\selectfont
\centering
\begin{tabular}{l|ccc|c}
\hline
\multirow{2}{*}{Method}          
& \multicolumn{3}{c|}{CIFAR10 N}    & \multicolumn{1}{c}{CIFAR100 N}       \\ 

& Aggre &Rand1  &Worst &Noisy Fine  \\ 
\hline
CE  &87.77 &85.02 &77.69 &55.50     \\
CoTeaching+  &91.20 &90.33 &83.83 &60.37     \\
JoCoR   &91.44 &90.30 &83.37 &59.97    \\
DivideMix  &95.01 &95.16 &92.56 &71.13     \\
ELR+  &94.83 &94.43 &91.09 &66.72     \\
CORES*  &95.25 &94.45 &91.66 &55.72     \\
SOP+  &95.61 &95.28 &93.24 &67.81     \\
PES(Semi)  &94.66 &95.06 &92.68 &70.36     \\
Promix  &\textbf{97.50} &97.23 &96.02 &73.62     \\
Promix+Ours  &97.49 &\textbf{97.29} &\textbf{96.16} &\textbf{73.66}     \\
\hline
\end{tabular}
\label{tab:cifarn}
\end{table}

\section{Semi-Supervised Learning for Label Denoising}
There is another popular paradigm for label noise learning, which first selects the clean sample and then trains the classifier by semi-supervised learning algorithms.
Here, we want to show that our method can also adapt to such a paradigm.
\paragraph{Problem Setup for Semi-Supervised Learning.}
Considering a classification problem with $k$ classes.
After sample selection, we obtain training examples divided into two sets: a clean set and a noisy set. Our objective remains to train a clean classifier that minimizes the empirical risk of $x$ on the clean label $y$ in the test set.

\begin{figure}[h]
     \centering
        \includegraphics[width=0.8\linewidth]{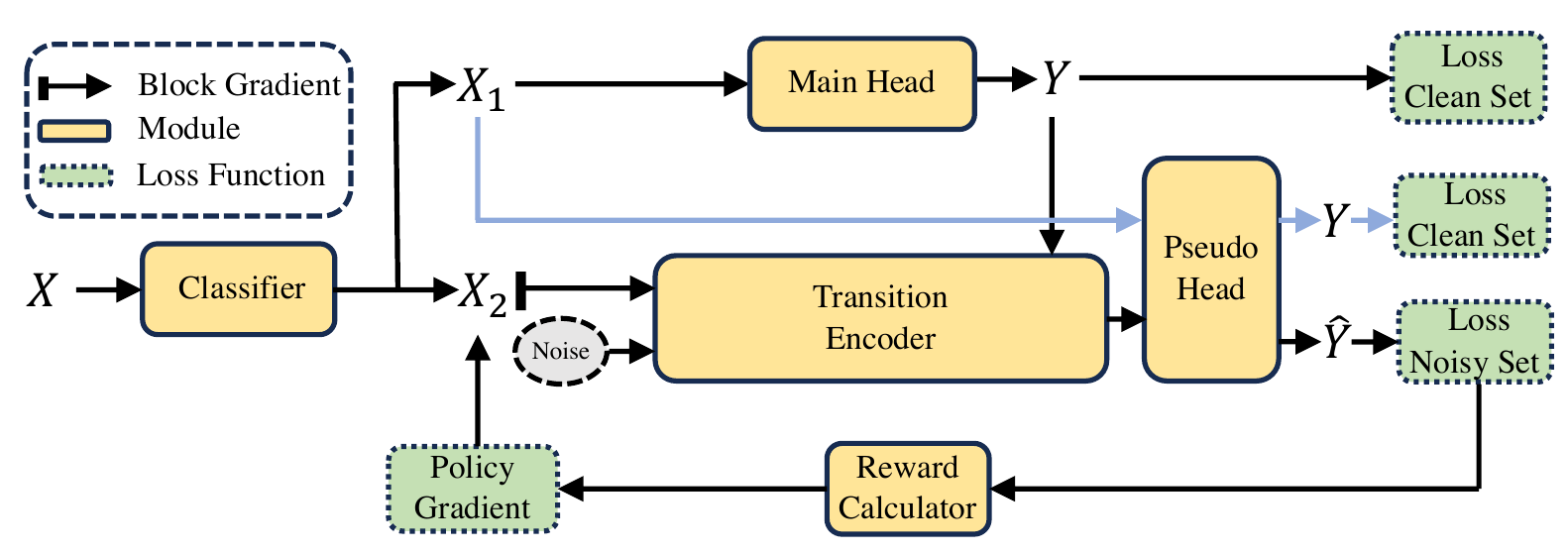}
  \caption{Denoising training framework in Semi-Supervised setting }
  \vspace{-1em}
    \label{fig:semi_framework}
\end{figure}

In this paradigm, two key steps need to be performed:

(i) Decorrelating $X_1$ and $\hat{Y}$.

(ii) Modeling the relationship from $X_2$, $Y$, $Z$ to $\hat{Y}$.

Given that we have access to some clean samples, we propose a more effective training framework for such settings, as illustrated in Figure~\ref{fig:semi_framework}.
First, we build two linear layers with activation functions to obtain $X_1$ and $X_2$. 
For $X_1$, we set up a main head to generate logits for the samples in the clean set.

Simultaneously, we set up a transition encoder that takes
$X$, $Y$, and $X_2$ as inputs and outputs a tensor with the same shape as $X_2$. This tensor is then fed into a pseudo head to produce logits for the samples in the noisy set. The transition encoder and pseudo-head together serve as the transition matrix, accomplishing step (ii).
To achieve step (i), we also feed
$X_1$ of the clean samples into the pseudo head but require it to output $Y$, making $X_1$ noise resistant.
Furthermore, we continue to update $X_2$ using the policy gradient.
During inference, only the main head is utilized.
The overall loss function is defined as:

\begin{equation}
\mathcal{L} = \mathcal{L}_{\text{clean}} + \alpha_1 \cdot \mathcal{L}_{\text{noisy}} + \alpha_2 \cdot \mathcal{L}_{PG} + \alpha_3 \cdot \mathcal{L}_{\text{Dec}}(X_1),
\label{eq}
\end{equation}
where 
 $\mathcal{L}_{\text{clean}}$ 
 represents the loss on the clean set, 
$\mathcal{L}_{\text{noisy}}$
 represents the loss on the noisy set, 
$\mathcal{L}_{PG}$
 represents the policy gradient loss, 
 $\mathcal{L}_{\text{Dec}}$
 represents the decorrelation loss for 
$X_1$
and $\alpha_1$, $\alpha_2$, $\alpha_3$ denotes the scale factors.
We implemented our method based on Promix~\cite{xiao2023promix}, one of the state-of-the-art method that selects clean examples with high confidence
scores and dynamically expand a base clean sample set. The results on the CIFAR-N dataset\footnote{We adopt the experiment results of the baselines from ~\citet{xiao2023promix} and implement Promix and our proposed method on an NVIDIA V100 GPU.}~\cite{wei2021learning} are listed in Table~\ref{tab:cifarn}.
We set $\alpha_1=1$ in all scenarios. Additionally, we set $\alpha_2=0.1$ and $\alpha_3=0.1$ for CIFAR-10N, and $\alpha_2=1$  and $\alpha_3=0.01$ for CIFAR-100N.

\section{Limitations}
The primary limitation of our method is that it depends on the assumption that the clean label $Y$ is partially observable, which is facilitated by confidence sampling. However, in scenarios where there is a high rate of label noise, obtaining the clean label $Y$ can be challenging, potentially resulting in an unwanted performance of our approach.

\end{document}